\documentclass[letterpaper]{article} 
\usepackage{aaai2027}  
\usepackage[hyphens]{url}  
\usepackage{graphicx} 
\usepackage{natbib}  
\usepackage{caption} 
\usepackage{algorithm}
\usepackage{algorithmic}

\usepackage{amsmath}
\usepackage{amssymb}
\usepackage[table]{xcolor}
\usepackage{siunitx}
\usepackage{newfloat}
\usepackage{listings}
\DeclareCaptionStyle{ruled}{labelfont=normalfont,labelsep=colon,strut=off} 
\floatstyle{ruled}
\newfloat{listing}{tb}{lst}{}
\floatname{listing}{Listing}

\usepackage{booktabs}

\title{RLPF: Reinforcement Learning from Performance Feedback for Code Generation}
\newcommand{\hku}{%
\includegraphics[height=1em]{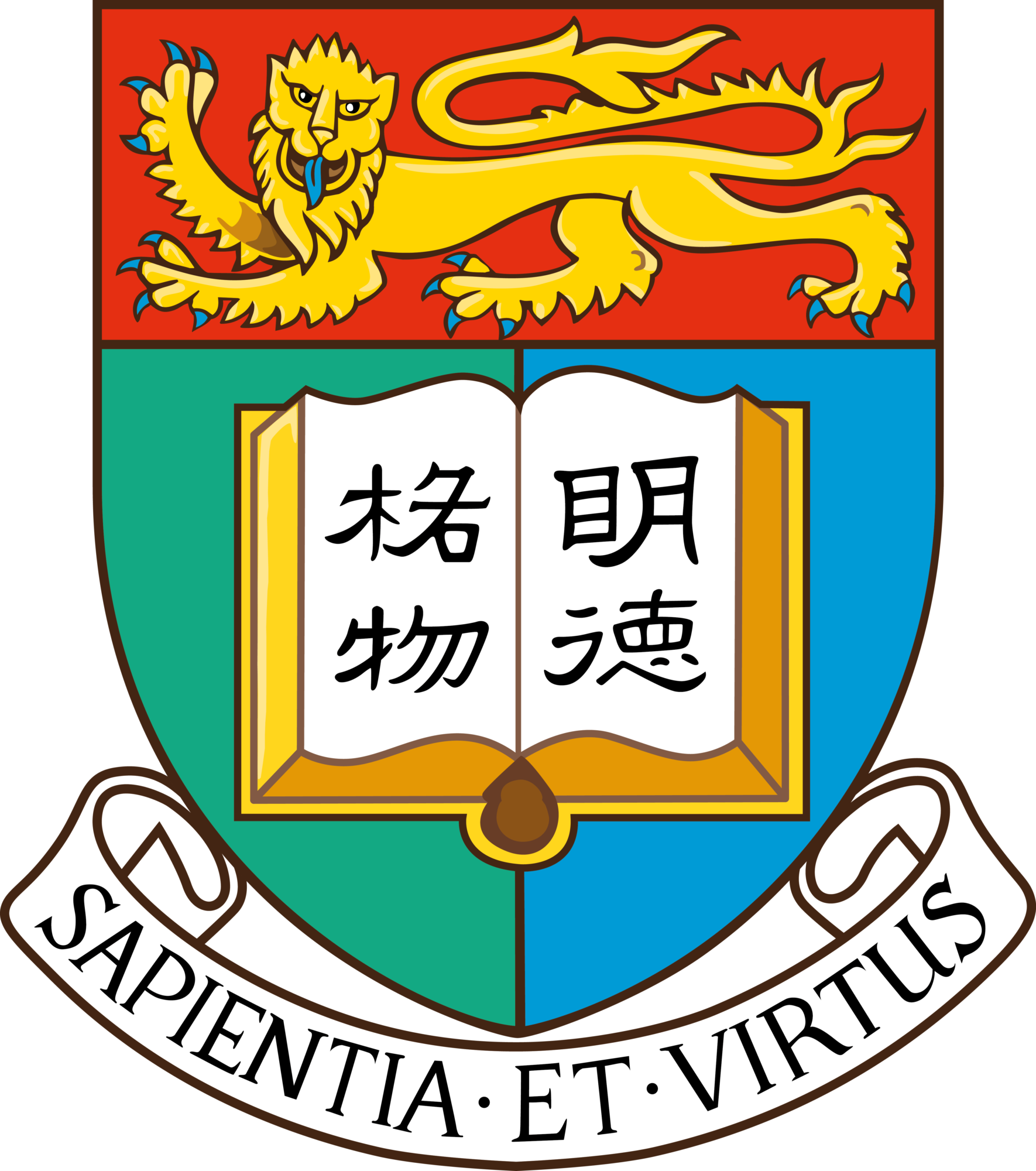}%
}
\newcommand{\hkusticon}{%
\includegraphics[height=1em]{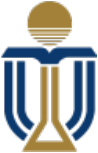}%
}
\newcommand{\nyuicon}{%
\includegraphics[height=1em]{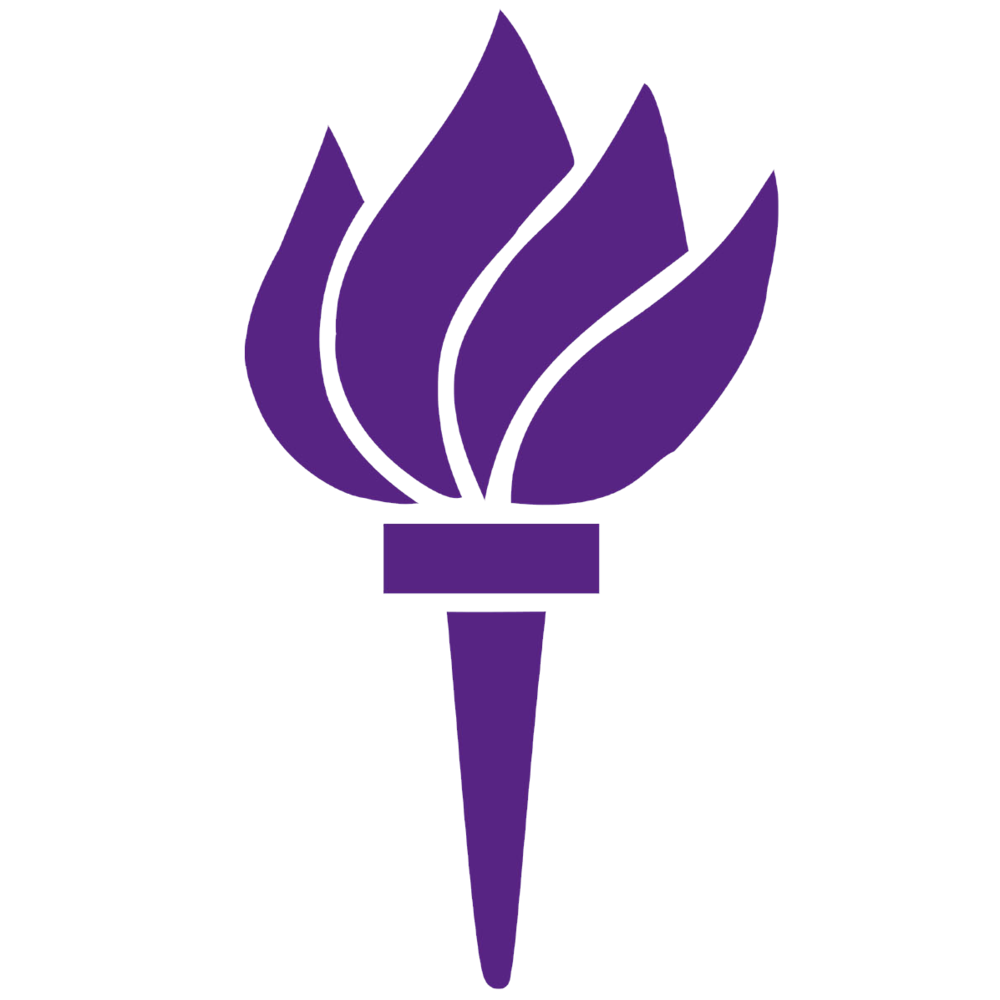}%
}
\newcommand{\swuplicon}{%
\includegraphics[height=1em]{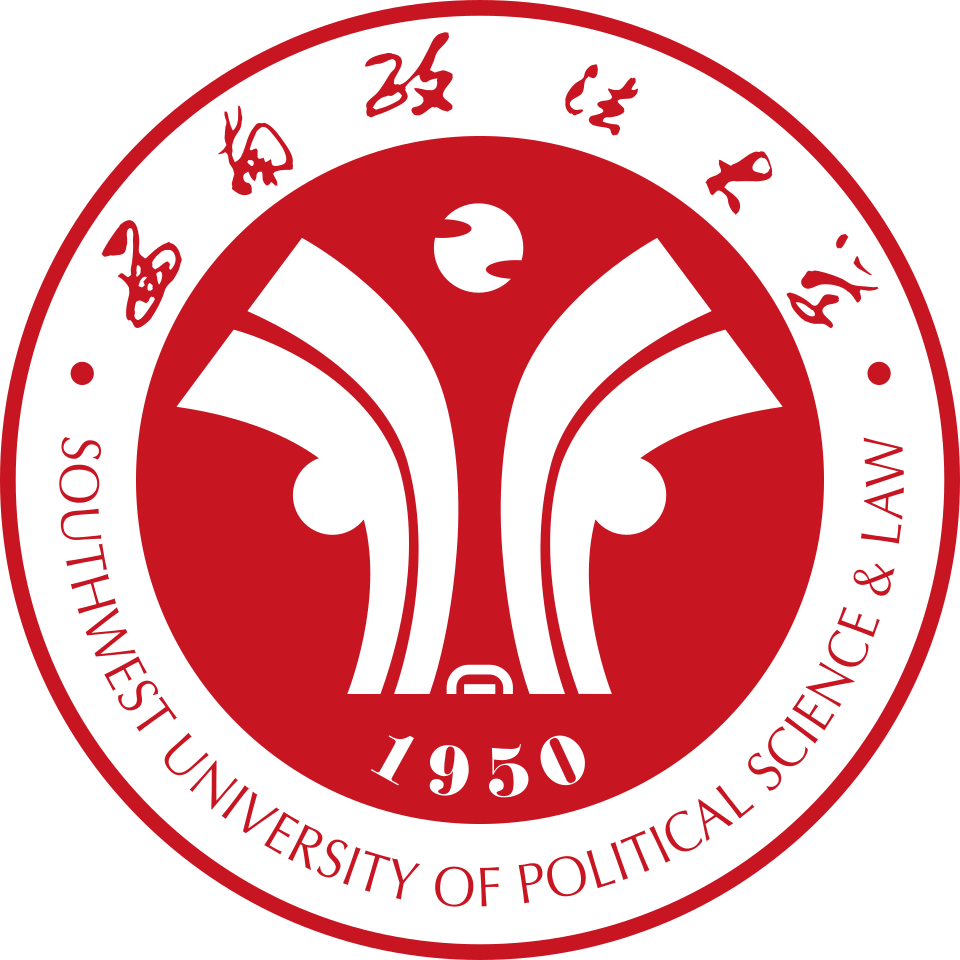}%
}
\newcommand{\modeioicon}{%
\includegraphics[height=1em]{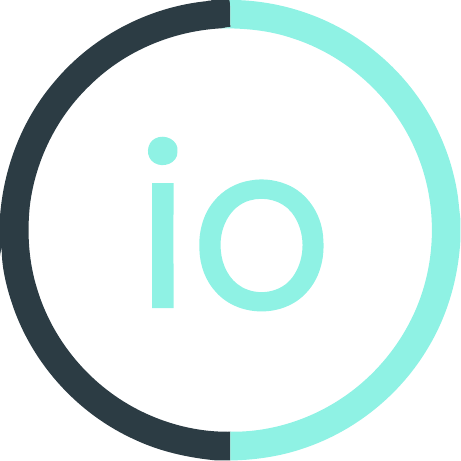}%
}

\author{
Huihao Jing\textsuperscript{\hkusticon},
Haozhe Cui\textsuperscript{\hku},
Wenbin Hu\textsuperscript{\hkusticon},
Shaojin Chen\textsuperscript{\hkusticon},
Haochen Shi\textsuperscript{\hkusticon},
Changxuan Fan\textsuperscript{\hkusticon}, \\
Yuxuan Liu\textsuperscript{\hkusticon},
Hanyu Yang\textsuperscript{\swuplicon\hspace{0.1em}\modeioicon},
Sirui Zhang\textsuperscript{\nyuicon\hspace{0.1em}\modeioicon},
Ziyi Chen\textsuperscript{\modeioicon},
Haoran Li\textsuperscript{\hkusticon}\corresponding,
Yangqiu Song\textsuperscript{\hkusticon}
}

\affiliations{
\textsuperscript{\hkusticon}%
The Hong Kong University of Science and Technology (HKUST),\quad
\textsuperscript{\nyuicon}%
New York University (NYU),\\
\textsuperscript{\swuplicon}%
Southwest University of Political Science and Law (SWUPL),\quad
\textsuperscript{\hku}%
Hong Kong University,\quad
\textsuperscript{\modeioicon}%
MODEIO.AI\\
hjingaa@connect.ust.hk
}

\begin{document}

\maketitle

\begin{abstract}
Code models are increasingly trained with execution feedback, but most training signals still stop at correctness. This leaves an important gap for systems code: two programs can pass the same tests while differing greatly in runtime. We study how to train code agents to prefer faster correct implementations, rather than treating efficiency only as an evaluation metric. The key difficulty is that runtime is a fragile reward. It is meaningful only after a program is correct, varies across tasks, and gives little guidance when most sampled programs fail to compile or run. We propose \textbf{RLPF}, reinforcement learning from performance feedback, which turns execution outcomes into a staged reward. Failed programs are ordered by execution progress, while correct programs are ranked by their relative improvement from the baseline toward the expert reference. This gives useful feedback before correctness and performance-sensitive feedback after correctness. Fine-tuning Qwen3-32B with RLPF on PerfCodeBench raises correct-and-runnable solutions from $11.1\%$ to $54.6\%$ and improves relative efficiency from $8.1\%$ to $38.6\%$. The trained model becomes competitive with stronger open-weight systems, and its optimization behavior transfers modestly to EffiBench-X. Additional studies show that model-generated references provide useful but weaker supervision, and that the full composite reward is more reliable than correctness-only or runtime-only baselines. These results suggest that code agents can be trained not only to pass tests, but also to optimize the programs they write.
\end{abstract}

\begin{links}
    \link{Code}{https://github.com/HKUST-KnowComp/RLPF}
    \link{Models}{https://huggingface.co/Egbertjing/RLPF-Qwen3-32B-PerfCodeBench}
\end{links}

\section{Introduction}
\label{sec:intro}

Large language models (LLMs) have become capable code generators. Frontier systems such as Claude Code~\cite{anthropic2026claudecode} and Codex agents~\cite{openai2026codex}, together with a growing line of agentic software-engineering research~\cite{DBLP:conf/nips/YangJWLYNP24,DBLP:conf/iclr/0001LSXTZPSLSTL25}, can navigate repositories, run tests, and repair their own outputs. Correctness-oriented benchmarks, from function-level synthesis to repository-scale issue resolution~\cite{DBLP:conf/iclr/JimenezYWYPPN24,DBLP:conf/iclr/JainHGLYZWSSS25,DBLP:conf/iclr/ZhuoVCH0WYZHPB025}, have therefore made executable correctness the central measure of code-model progress. For performance-critical code, this is only a partial goal. A program that passes all tests may still be unusable if it is much slower than a simple baseline or far from an optimized implementation. Recent efficiency-oriented studies~\cite{DBLP:conf/nips/0005QSCZ24,du2024mercury,qiu2024enamel,liu2024evalperf,DBLP:journals/corr/abs-2505-13004,ouyang2025kernelbench,peng2025coffe,perfcodebench2026} show this gap clearly: strong models can often produce correct programs, but their solutions still fail to match expert runtimes. As code agents are used in more realistic software workflows, the question is no longer only whether they can write working code. It is whether they can learn to prefer better working code. This makes performance a natural target for reinforcement learning, but not a simple one. Runtime is an execution-based signal. It can be observed only after the generated program is valid, runs, and passes the oracle. It also has task-dependent scale: a one-second improvement on one problem and a one-millisecond improvement on another do not carry the same meaning. A naive runtime or speedup reward therefore gives unstable supervision across heterogeneous tasks. The problem is sharper under group-relative policy optimization (GRPO)~\cite{shao2024deepseekmath}. If every rollout in a group fails before correctness, runtime cannot provide a useful ordering, and the policy receives little signal exactly when training is hardest.

We propose \textbf{RLPF}, reinforcement learning from performance feedback. The main idea is to reward the execution process in stages. Before correctness, RLPF orders failed programs by how far they progress through the executable pipeline, from extraction and compilation to execution and oracle checking. After correctness, it ranks programs by how much of the baseline-to-reference performance gap they close. This design gives the policy two kinds of feedback in one training run: progress feedback for failed programs, and efficiency feedback for correct programs. It also makes performance comparable across tasks by measuring improvement relative to the task's own baseline and expert reference.

We instantiate RLPF on PerfCodeBench. On the family-disjoint test split, RLPF raises a weak base model from $11.1\%$ to $54.6\%$ correct-and-runnable rate and from $8.1\%$ to $38.6\%$ CGRE. The improvement is not merely correctness inflation: most correct outputs also beat the baseline. The trained model becomes competitive with stronger open-weight systems, while a clear gap to expert references and proprietary frontier models remains. We further test a weaker, distillation-like setting where GPT-5.4 outputs serve as performance references; these references help, but are less stable than curated expert implementations. Finally, reward baselines, component ablations, and EffiBench-X transfer results show that the full reward is more reliable than correctness-only or runtime-only supervision and that the learned preference for faster correct code partly transfers beyond the training benchmark.

Our contributions are as follows:
\begin{itemize}
  \item \textbf{We formulate performance as a trainable objective for code agents.} RLPF moves beyond test passing by rewarding correct programs according to their efficiency relative to a baseline and an expert reference.
  \item \textbf{We introduce a staged reward for heterogeneous execution outcomes.} The reward provides progress feedback before correctness and performance feedback after correctness, avoiding the sparse signal of naive runtime rewards.
  \item \textbf{We show that performance feedback changes model behavior.} RLPF substantially improves Qwen3-32B on PerfCodeBench, transfers modestly to EffiBench-X, and controlled studies identify both the value and the limits of model-generated references, simple reward baselines, and reward components.
\end{itemize}

\section{Related Work}
\label{sec:related}

\subsection{LLM Code Generation.} 
Large language models have made rapid progress in code generation, from function-level synthesis to competitive programming and software-engineering workflows. This progress has been shaped by executable benchmarks such as HumanEval and EvalPlus~\cite{chen2021evaluating,liu2023evalplus}, MBPP~\cite{austin2021program}, APPS~\cite{hendrycks2021measuring}, LiveCodeBench~\cite{DBLP:conf/iclr/JainHGLYZWSSS25}, and SWE-bench~\cite{DBLP:conf/iclr/JimenezYWYPPN24}, which evaluate generated code through unit tests, execution, or realistic repository-level tasks. Recent benchmarks further move toward practical developer settings; DevBench~\cite{kumarappan2026devbench}, for example, is a telemetry-driven and developer-informed benchmark covering realistic code-completion scenarios. As a result, functional correctness has become the default measure of code quality, and much follow-up work focuses on improving test passing through execution feedback, self-debugging, retrieval, multi-agent collaboration, or test-time scaling. However, passing tests is not enough for performance-critical code: two correct programs can differ greatly in runtime, memory use, and hardware efficiency. PIE, PerfCodeGen, EffiPair, and PhyloEvolve use execution feedback or paired edits to produce faster implementations~\cite{shypula2023pie,peng2024perfcodegen,hajizadeh2026effipair,zhao2026phyloevolve}. Efficiency benchmarks now cover general-purpose languages and GPU kernels~\cite{du2024mercury,qiu2024enamel,liu2024evalperf,DBLP:journals/corr/abs-2505-13004,ouyang2025kernelbench,perfcodebench2026}. EffiCoder shows that SFT on curated optimized solutions can improve correctness and efficiency~\cite{huang2025efficoder}. RLPF instead uses online RL, combining failure-stage feedback with correctness-gated, task-normalized performance rewards.

\subsection{Reinforcement Learning for Code.} 
Reinforcement learning has recently emerged as an important post-training paradigm for code LLMs, especially through reinforcement learning with verifiable rewards (RLVR)~\cite{le2022coderl,liu2023rltf,yu2024bcoder,shen2023pangucoder2,deepseek2025r1}. In code generation, verifiable rewards are typically derived from execution results, allowing models to optimize directly against executable correctness signals. Recent work extends this paradigm in several directions. RLEF trains models to use execution feedback during code synthesis~\cite{gehring2025rlef}. CodeRL+ augments sparse pass/fail rewards with execution-semantics alignment~\cite{jiang2025coderlplus}. CodeScaler uses learned reward models to reduce reliance on online execution~\cite{zhu2026codescaler}. Other studies explore offline RL~\cite{wu2026offlinecode}, verification feedback for small models~\cite{skopin2026verification}, collaborative multi-agent RL~\cite{dou2026corecode}, and synthetic-data-driven code RL~\cite{wu2026xcoder}.

Despite this progress, code RL remains largely correctness-centric: rewards usually separate correct programs from incorrect ones, while efficiency is left as a post-training evaluation metric. Such training improves test passing, but does not teach models to prefer faster correct implementations. RLPF addresses this gap by treating efficiency as a first-class verifiable signal, extending RL for code from verifiable correctness to verified optimization.

\subsection{Speedup Rewards for Code Optimization.} 
A natural way to add performance feedback is to use runtime improvement, speedup, throughput, or hardware utilization as the reward. Several recent optimization-oriented systems follow this idea. Wei et al.~\cite{wei2025assembly} study RL for assembly code optimization and compare correctness-guided speedup with a speedup-only reward. CUDA-L1~\cite{li2025cudal1} trains LLMs for CUDA kernel optimization with speedup-based rewards. Mikasa et al.~\cite{mikasa2026hpc} use real-machine GFLOPS feedback as the reward for online GRPO in HPC code generation. MaxCode~\cite{ou2026maxcode} similarly treats execution performance as the central signal during candidate search.

These works show that performance feedback can improve code optimization. However, they usually use the final measured performance of a candidate program, such as speedup, runtime, throughput, or GFLOPS, as the main reward signal. This is effective when tasks are specialized and comparable, but it becomes unstable for general code generation. Different tasks have different runtime scales, optimization margins, and failure modes, so raw performance values are difficult to compare across problems. More importantly, they provide little useful signal when a program fails to compile, execute, or pass tests. RLPF therefore uses performance feedback only after building a denser execution-state signal. It first shapes compilation, execution, and correctness outcomes, and then ranks correct programs by relative efficiency. This preserves the benefit of speedup-based optimization while making the reward more stable across heterogeneous execution outcomes.

\section{Train LLMs with RLPF} 
\label{sec:method}

\begin{figure*}[t]
\centering
\includegraphics[width=\linewidth]{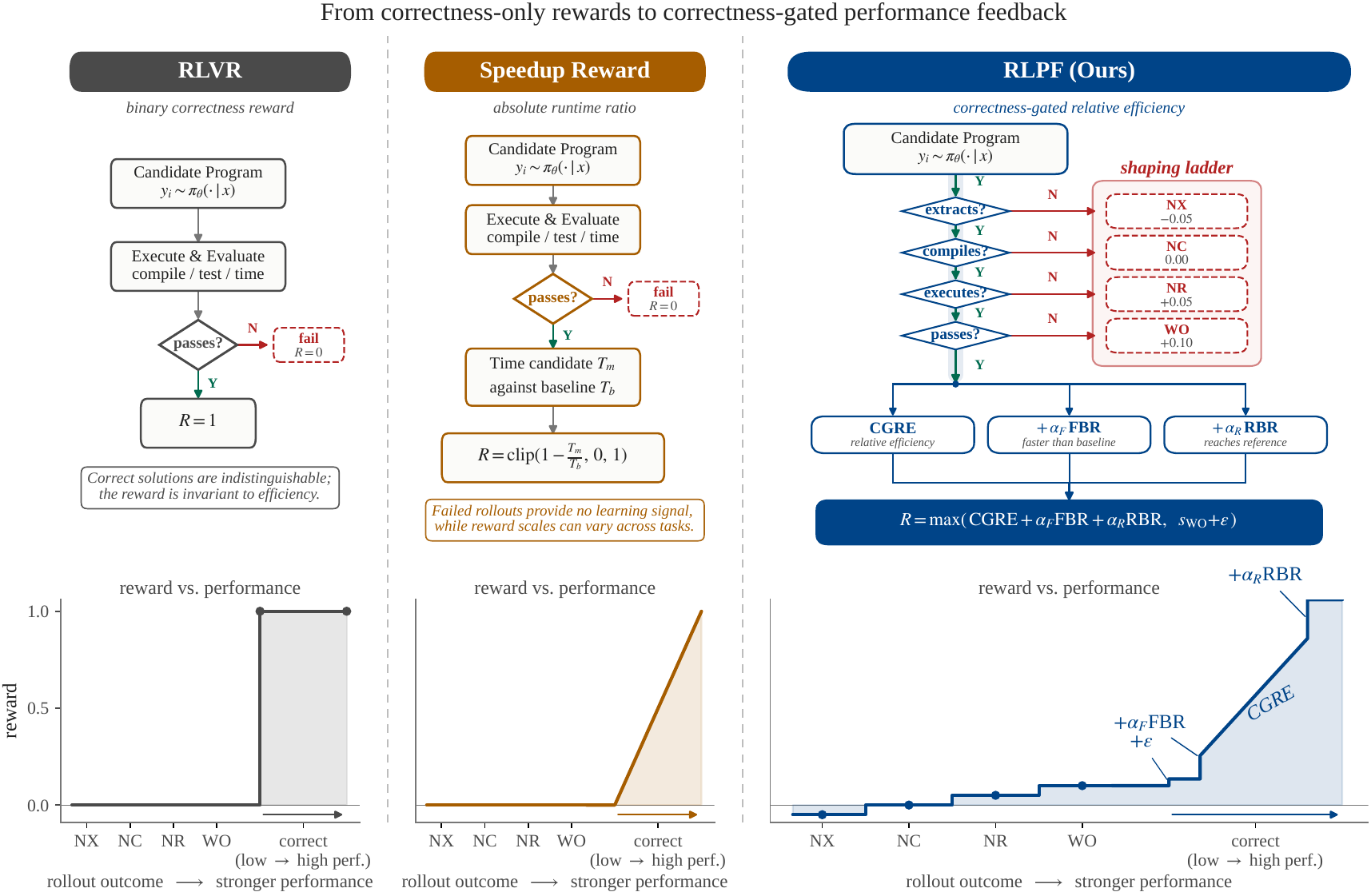}
\vspace{-0.20in}
\caption{\textbf{From correctness-only rewards to correctness-gated performance feedback.} RLVR rewards correctness alone and cannot distinguish fast correct programs from slow correct programs. A naive speedup reward uses runtime improvement as the main signal, but provides little feedback before correctness. RLPF decomposes execution outcomes into an ordered reward structure: failed rollouts receive staged shaping feedback, while correct rollouts are ranked by correctness-gated relative efficiency.}
\label{fig:reward-design}
\vspace{-0.1in}
\end{figure*}

RLPF is based on the principle that code-performance training should reward the entire execution process, rather than treating performance as a single terminal scalar. A generated program must first be extractable, compilable, executable, and correct before runtime efficiency becomes meaningful. Therefore, RLPF separates reward design into two regimes. In the \emph{failure regime}, the reward measures execution progress: how far an incorrect rollout moves through the execution pipeline. In the \emph{success regime}, the reward measures relative performance: how efficient a correct rollout is compared with the task baseline and expert reference. This design lets RLPF provide learning signal even before correctness, while still optimizing performance once correctness is achieved.

We optimize the policy with group-relative policy optimization (GRPO)~\cite{schulman2017ppo,shao2024deepseekmath}. GRPO compares multiple rollouts from the same prompt, so the reward must induce a meaningful ordering among candidate programs. In RLPF, this ordering is defined by execution outcomes rather than by a single runtime scalar. Failed rollouts are ranked by how far they progress through the execution pipeline, while correct rollouts are ranked by their performance improvement relative to the task baseline and expert reference. This lets the same reward structure cover the full outcome space, from invalid generations to optimized correct programs.

\subsection{Problem Setup}

We consider performance-oriented code generation. Each task $i$ provides a natural-language problem description, a required function signature, a baseline implementation, an expert reference implementation, and an executable test harness. The baseline implementation defines a valid but unoptimized starting point, while the expert reference is a stronger implementation written by human experts and serves as the target performance level. Given the problem description, the model generates a candidate program. The harness then evaluates the candidate through the execution pipeline: it extracts the generated code, compiles it when needed, executes it, checks oracle tests, and measures runtime. If the candidate passes the tests, its runtime is compared with both the baseline and the expert reference to measure the model’s performance-improvement ability.

\subsection{Success-Mode Reward: Relative Efficiency}

For task $i$, let $T_b^i$, $T_r^i$, and $T_c^i$ denote the median wall-clock runtime of the baseline implementation, the expert reference, and the candidate program. Let $C_i\in\{0,1\}$ indicate whether the candidate passes all oracle tests. For correct programs, we score efficiency using three signals:
\begin{align}
\mathrm{FBR}_i
&= C_i \cdot \mathbf{1}\!\left[T_c^i < T_b^i\right], \\
\mathrm{RBR}_i
&= C_i \cdot \mathbf{1}\!\left[T_c^i \le T_r^i\right], \\
\mathrm{CGRE}_i
&= C_i \cdot
\operatorname{clip}\!\left(
\frac{T_b^i - T_c^i}{T_b^i - T_r^i},\,0,\,1
\right).
\label{eq:efficiency-metrics}
\end{align}

FBR indicates whether the candidate is faster than the baseline. RBR indicates whether it reaches or beats the expert reference. CGRE is the dominant continuous term: it measures how much of the baseline-to-reference performance gap the candidate closes. Unlike raw runtime or raw speedup, CGRE is normalized by the task-specific optimization gap, making it more comparable across tasks with different runtime scales and optimization margins.

The success-mode reward is
\begin{equation}
\begin{aligned}
R_i^{\mathrm{succ}}
&= \mathrm{CGRE}_i
+ \alpha_F \mathrm{FBR}_i
+ \alpha_R \mathrm{RBR}_i .
    \end{aligned}
    \label{eq:rsucc}
    \end{equation}
We use $\alpha_F=0.16$ and $\alpha_R=0.10$. CGRE remains the main efficiency signal, while FBR and RBR provide small bonuses for crossing important performance thresholds.

\subsection{Failure-Mode Reward: Execution Progress}

The success-mode reward is only meaningful for correct programs. However, in early training, especially on hard systems tasks, most rollouts may fail before correctness. If failed rollouts are not distinguished, GRPO receives an overly sparse reward signal: most early samples collapse to the same failure value, making it hard for the policy to learn which generations are closer to correctness.

RLPF therefore assigns failed rollouts a separate shaping reward based on their execution state. We use four pre-correct failure modes, $\sigma\in{\mathrm{NX},\mathrm{NC},\mathrm{NR},\mathrm{WO}}$, where NX means no extractable program, NC means no compilation, NR means no successful execution, and WO means wrong output after execution. These states form a natural progress order: producing parseable code is better than producing no code; compiling is better than failing to compile; running is better than crashing; and producing a wrong answer is closer to success than failing earlier. We encode this order as
\begin{equation}
s(\sigma)=
\begin{cases}
-0.05, & \sigma = \mathrm{NX},\\
\phantom{-}0.00, & \sigma = \mathrm{NC},\\
+0.05, & \sigma = \mathrm{NR},\\
+0.10, & \sigma = \mathrm{WO}.
\end{cases}
\label{eq:failure-shaping}
\end{equation}

This failure-mode reward does not optimize speed directly. Instead, it teaches the model to climb the execution pipeline until correctness becomes reachable.

\subsection{Final Reward Staircase}

The final reward combines the two regimes with an explicit gap between failure and success. For failed rollouts, RLPF uses the failure-mode shaping reward directly. For correct rollouts, RLPF uses the larger value between the success-mode efficiency reward and a correctness floor:
\begin{equation}
R_i =
\begin{cases}
s(\sigma_i),
& C_i = 0,\\
\max\!\left\{
s(\mathrm{WO})+\epsilon,\,
R_i^{\mathrm{succ}}
\right\},
& C_i = 1.
\end{cases}
\label{eq:rfinal}
\end{equation}
Here $s(\mathrm{WO})$ is the highest pre-correct reward, and $\epsilon>0$ creates a strict gap between the best failed rollout and the worst correct rollout. In our implementation, $s(\mathrm{WO})=0.10$ and $\epsilon=0.05$, so every correct program receives at least $0.15$, even if it does not improve over the baseline.

This produces the following reward staircase:
\begin{equation}
\begin{aligned}
\mathrm{NX}
&< \mathrm{NC}
< \mathrm{NR}
< \mathrm{WO}
< \text{correct} \\
&< \text{faster-than-baseline}
< \text{reference-or-better}.
\end{aligned}
\label{eq:reward-staircase}
\end{equation}
The first part of the staircase ranks failed programs by execution progress, while the second part ranks correct programs by efficiency. Thus, RLPF provides dense feedback before correctness and performance-sensitive feedback after correctness, while ensuring that correctness remains the boundary between the two regimes.

\section{Experiment Setup} 
\label{sec:exp-setup}

\subsection{Benchmarks and Evaluation Protocol}

We use PerfCodeBench~\cite{perfcodebench2026} for both training and in-distribution evaluation. It matches the setting targeted by RLPF: each task asks the model to produce a drop-in implementation, which is then checked for correctness and compared with a baseline and an expert reference for efficiency. We train on the $1{,}413$ tasks in the PerfCodeBench \emph{train} split and evaluate on the family-disjoint \emph{test} split of $306$ tasks. The benchmark covers C, C++, CUDA, Go, Java, and Python. We also evaluate out of distribution on the C++, Python, Java, Go, and JavaScript subset of EffiBench-X~\cite{DBLP:journals/corr/abs-2505-13004}. No EffiBench-X data is used during training. Because it does not use the same baseline-reference setup as PerfCodeBench, we treat it as a transfer test: we report Pass@1 for correctness, and for efficiency we compare the trained model with the base model only on problems both solve, using geometric-mean improvement and win rate.

\subsection{Metrics}

For PerfCodeBench, we report CRR, FBR, RBR, CGRE, and $\mathrm{CGRE}_{\ge 0.8}$. The last metric is the percentage of all test tasks---not only the correctly solved subset---on which a correct candidate closes at least $80\%$ of the baseline-to-reference performance gap, i.e., $\mathrm{CGRE}_i\ge 0.8$. Together, these metrics separate correctness, improvement over the baseline, expert-level performance, and the fraction of the expert gap that is closed.

For EffiBench-X, we use a separate protocol because it does not provide the same baseline-reference setup as PerfCodeBench. For each shared problem $k$, let $\rho_k$ be the ratio between the base-model measurement and the trained-model measurement, so $\rho_k>1$ means the trained model improves over the base model. We summarize the paired improvement by the geometric mean:
\begin{equation}
\Delta = \exp\!\left(\frac{1}{N}\sum_{k=1}^{N} \ln \rho_k\right) - 1,
\label{eq:paired-gain}
\end{equation}
We report $\Delta$ as a percentage for execution time (ET) and memory integral (MI), together with the ET win rate, i.e., the fraction of shared problems on which the trained model is faster than the base model.

\subsection{Training Setup}

Our policy model is \textsc{Qwen3-32B}~\cite{yang2025qwen3}, fine-tuned with GRPO~\cite{shao2024deepseekmath} and LoRA~\cite{hu2022lora} under the RLPF reward. For each prompt, the policy samples $G=8$ candidate programs, which are then scored by the same executable harness used in evaluation. The reward is the composite objective in Eqs.~\ref{eq:rsucc}--\ref{eq:rfinal}, with $\alpha_F=0.16$ and $\alpha_R=0.10$. We keep the base model frozen. The full RLPF model, reward baselines, and ablation variants are each trained for five epochs. During training, runtime is measured once; during evaluation, we report the median of $K=3$ runs.

\subsection{Baseline and Ablation Setup}

We use the same model, training split, decoding setting, and evaluation harness for all reward baselines and ablations. The goal is to isolate the training signal, so we do not change the base model or the test protocol across variants. We first include two simple reward baselines. The runtime-only baseline replaces the full reward with a naive runtime signal. The RLVR baseline uses a binary correctness reward: a rollout receives reward only for passing the oracle, with no performance term and no failure-mode shaping. We then run component ablations that remove one part of the composite reward at a time: the faster-than-baseline bonus, the reference-or-better bonus, or the failure-mode shaping ladder. Together, these runs test whether RLPF works because of one dominant term or because the terms provide complementary feedback.

For the model-generated reference experiment, we keep the RLPF reward structure but replace the expert reference target with a verified GPT-5.4 candidate when one is available. The teacher candidate must pass the same correctness oracle before its measured runtime is used as the reference. This variant tests a weaker data setting, where performance supervision comes from a strong model rather than from a curated expert implementation.

\subsection{Checkpoint Selection}

For every model, including the full RLPF recipe, reward baselines, and ablation variants, we select the checkpoint at the exponential-moving-average (EMA) peak of its training reward curve. This unified criterion compares each variant near its own best observed training point rather than at an arbitrary step. All models are decoded with thinking disabled, matching the training regime. No EffiBench-X data is used during training.

\section{Experiment Results}
\label{sec:exp-main}

\begin{table*}[t]
\centering
\setlength{\tabcolsep}{5.6pt}
\renewcommand{\arraystretch}{1.08}
\sisetup{
    detect-weight=true,
    detect-family=true,
    table-number-alignment=center,
    round-mode=places,
    round-precision=2
}

\begin{tabular}{
    l
    S[table-format=2.2]
    S[table-format=2.2]
    S[table-format=2.2]
    S[table-format=2.2]
    S[table-format=2.2]
    S[table-format=2.2]
    S[table-format=2.2]
}
\toprule
\textbf{Model}
& {\textbf{CRR}}
& {\textbf{FBR}}
& {\textbf{RBR}}
& {\textbf{Slow/CRR}}
& {\textbf{Gap/FBR}}
& {\textbf{CGRE}}
& {\textbf{CGRE$\geq$0.8}} \\
\midrule

\multicolumn{8}{l}{\textit{\textbf{Ours: Qwen3-32B + GRPO on PerfCodeBench}}} \\
\rowcolor{blue!6}
\textsc{RLPF-32B}
& 54.58 & 46.41 & 25.82 & 14.97 & 44.37 & 38.58 & 37.25 \\
\rowcolor{blue!3}
\quad RLPF w. GPT-5.4 reference
& 50.33 & 42.81 & 25.49 & 14.94 & 40.46 & 36.53 & 35.29 \\
\rowcolor{blue!3}
\quad w/o FBR
& 40.52 & 32.03 & 18.30 & 20.95 & 42.87 & 25.44 & 23.20 \\
\rowcolor{blue!3}
\quad w/o RBR
& 39.87 & 31.05 & 19.61 & 22.12 & 36.84 & 25.14 & 23.53 \\
\rowcolor{blue!3}
\quad w/o shaping
& 39.54 & 31.05 & 18.95 & 21.47 & 38.97 & 24.22 & 22.22 \\
\rowcolor{blue!3}
Runtime reward only
& 37.91 & 31.70 & 17.65 & 16.38 & 44.32 & 24.66 & 22.55 \\
\rowcolor{blue!3}
RLVR correctness only
& 50.00 & 31.70 & 16.34 & 36.60 & 48.45 & 28.98 & 28.43 \\
Qwen3-32B (base, no RL)
& 11.11 & 8.17 & 5.56 & 26.46 & 31.95 & 8.11 & 8.17 \\

\addlinespace[2pt]
\midrule
\multicolumn{8}{l}{\textit{\textbf{Frontier and strong general-purpose models}}} \\
GPT-5.4
& \bfseries 78.10
& \bfseries 63.07
& \bfseries 50.65
& 19.24
& 19.69
& \bfseries 56.24
& \bfseries 55.23 \\
Claude Opus 4.5
& 65.03 & 53.59 & 39.87 & 17.59 & 25.60 & 45.59 & 43.79 \\
Gemini 3.1 Pro (preview)
& 34.64 & 30.07 & 23.20 & \bfseries 13.19 & 22.85 & 26.47 & 26.14 \\
DeepSeek-V4-Pro
& 64.38 & 55.56 & 49.02 & 13.70 & \bfseries 11.77 & 49.83 & 48.04 \\
Qwen3.6-Plus
& 46.73 & 38.24 & 21.90 & 18.17 & 42.73 & 32.93 & 31.70 \\
Qwen3.6-Max (preview)
& 43.14 & 35.29 & 16.67 & 18.20 & 52.76 & 30.40 & 29.74 \\
Gemma-4-26B-A4B-IT
& 59.80 & 45.42 & 26.47 & 24.05 & 41.72 & 39.65 & 37.25 \\
Gemma-4-31B-IT
& 49.67 & 37.25 & 18.63 & 25.01 & 49.99 & 32.87 & 32.35 \\
\bottomrule
\end{tabular}
\caption{
PerfCodeBench \emph{test} split leaderboard on 306 family-disjoint tasks.
All values are percentages. CRR is correct-and-runnable; FBR
(faster-than-baseline) and RBR (reference-or-better) are nested within CRR.
\textbf{Slow/CRR}\,$=$\,$(\mathrm{CRR}-\mathrm{FBR})/\mathrm{CRR}$ is the
percentage of correct solutions that are not faster than the baseline;
\textbf{Gap/FBR}\,$=$\,$(\mathrm{FBR}-\mathrm{RBR})/\mathrm{FBR}$ is the
percentage of faster-than-baseline solutions that still trail the expert
reference. Lower values are better for both conditional rates.
Our \textsc{RLPF} variants are highlighted, including one variant that uses a
verified GPT-5.4 candidate as the performance reference. Reward studies include
component removals and simple correctness-only or runtime-only baselines. Best
results are shown in \textbf{bold}.
}
\label{tab:perfcodebench-leaderboard}
\vspace{-0.22in}
\end{table*}

Table~\ref{tab:perfcodebench-leaderboard} reports the PerfCodeBench test-split leaderboard. We first compare the full \textsc{RLPF} recipe against the \textsc{Qwen3-32B} base model and against strong general-purpose systems.

\subsection{RLPF changes the failure regime}
The base \textsc{Qwen3-32B} model is a weak starting point for PerfCodeBench. It often fails before runtime can be measured, reaching only $11.1\%$ CRR. RLPF raises the same frozen base model with a LoRA adapter to $54.6\%$ CRR and $38.6$ CGRE. This is not a small score shift. It changes the dominant failure mode: the model moves from mostly failing the executable pipeline to producing usable implementations on a large part of the family-disjoint test split. This is important because performance feedback is only meaningful after several earlier steps succeed. A candidate must be extracted, compiled or loaded, run without crashing, and pass the oracle before its runtime can be interpreted. RLPF improves this whole path. It does not simply make already-correct programs faster; it moves many samples into the part of the search space where efficiency can be optimized at all. This helps explain why a small adapter can have a large effect on a difficult systems benchmark.

\subsection{Correctness gains are coupled with speed gains}
The improvement is not just more runnable code. The RLVR baseline is the clearest contrast: it reaches $50.0\%$ CRR, close to RLPF, but only $31.7\%$ FBR and $29.0$ CGRE. In other words, correctness-only training can teach the model to pass the harness, but many of those passing programs remain slow. RLPF has a similar correctness level but much higher efficiency, with roughly $85\%$ of its correct solutions beating the baseline. This coupling between correctness and speed is the main evidence that the reward changes the model's preference among correct implementations. The model learns that passing tests is not the endpoint. Among passing programs, it favors implementations that remove unnecessary work, use tighter loops, choose better primitives, or reduce data movement.

\subsection{Efficiency training improves model ranking}
RLPF also changes how the 32B model compares to stronger general-purpose systems. The trained model clearly surpasses its base model and outperforms Qwen3.6-Plus and Gemini~3.1~Pro on the main PerfCodeBench metrics. It is also close to Gemma-4-26B-A4B-IT: Gemma solves more tasks, but RLPF has slightly higher FBR and similar RBR. This suggests that performance-specific RL can recover part of the gap to stronger pretrained models, especially on tasks where the base model can already reach a valid implementation. The comparison also shows what RLPF does not solve. GPT-5.4, Claude Opus 4.5, and DeepSeek-V4-Pro remain clearly ahead, especially on RBR. These models not only produce more correct programs, but their correct programs also reach the expert reference more often. Thus RLPF improves the optimization policy of a weaker model, but it does not remove the value of broader pretrained competence, stronger code priors, and better task understanding.

\subsection{Baseline improvement and reference matching are different regimes}
The table separates two levels of efficiency. FBR measures whether a correct candidate beats the baseline. RBR measures whether it reaches the expert reference. RLPF improves both, but the gains are not equally strong. Its \textbf{Gap/FBR} rate is $44.4\%$: among solutions that beat the baseline, a substantial share still trails the expert reference. This difference is expected. Beating a baseline can often be achieved with local cleanup, less redundant work, better loop structure, or a simple library call. Matching the reference is harder. It may require a different algorithm, a specific memory layout, a parallel schedule, or careful use of the GPU or cache hierarchy. RLPF makes clear progress on the first regime and starts to improve the second, but reference-level optimization remains the main bottleneck.

\subsection{The metric breakdown reveals where progress is concentrated}
The metric columns are useful because they prevent a single leaderboard score from hiding the shape of the improvement. CRR shows whether the model reaches the correct-and-executable region. FBR shows whether those correct solutions are actually useful relative to the provided baseline. RBR and CGRE then distinguish partial speedups from expert-level ones. Under this breakdown, RLPF's strongest movement is from non-runnable or merely correct code toward faster-than-baseline code. Its weaker movement is from faster-than-baseline code to reference-or-better code. This gives a concrete diagnosis for future work: the reward already teaches the model to search for optimization opportunities, but more pressure or stronger initialization is needed near the expert ceiling.

\section{Ablation and Transfer Studies} 
\label{sec:exp-signal}

We next study why the method works and how far the learned behavior transfers. We vary the performance reference, ablate the reward components, and evaluate on an out-of-distribution benchmark.

\subsection{Model-Generated Performance References}

Expert references provide a clean performance target, but they also require strong benchmark construction. Each task needs a reliable optimized implementation, a stable harness, and enough runtime gap to define meaningful feedback. We therefore test a weaker data setting: can RLPF still work when the reference comes from a strong model rather than from an expert implementation? To study this, we first run GPT-5.4 on the training tasks. When its candidate is correct, we use that candidate's measured runtime as the reference target in the RLPF reward. This variant answers a practical question: can a frontier model provide useful performance supervision when expert references are unavailable? The answer is partly yes. The GPT-5.4-reference variant reaches $50.3\%$ CRR and $36.5\%$ CGRE, close to full RLPF, and it is also close on RBR. However, it still loses correctness and faster-than-baseline rate. Model-generated references are useful, but they are less uniform than expert references. They only help on tasks the teacher solves correctly, and the teacher's code may still miss task-specific optimizations.

\subsection{Reward Baselines and Component Ablations}
\label{sec:exp-ablation}

\begin{figure}[t]
\centering
\includegraphics[width=0.975\linewidth]{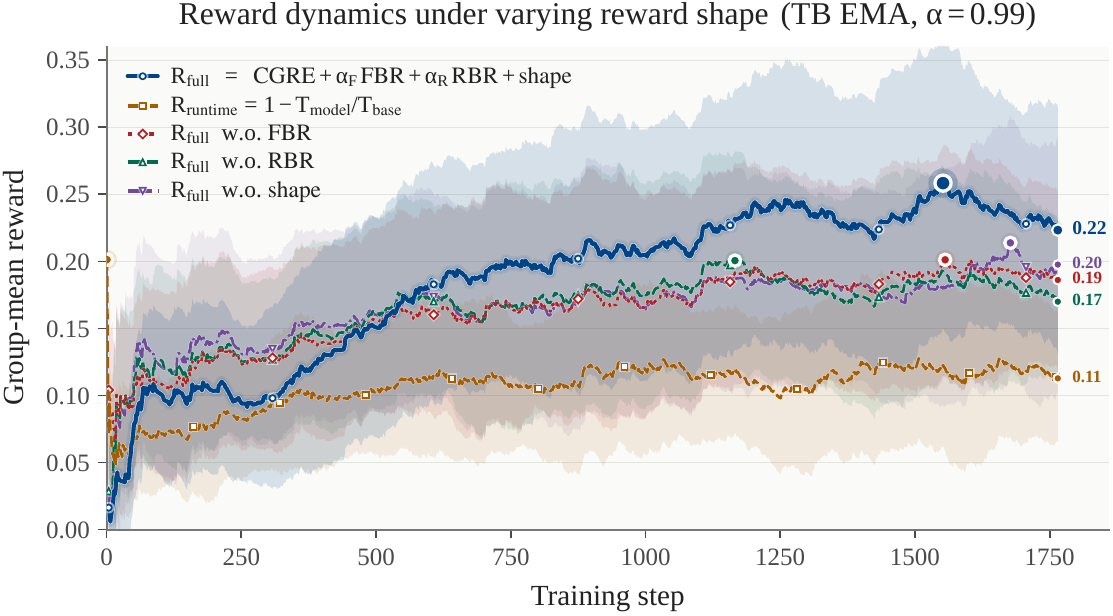}
\vspace{-0.05in}
\caption{Training dynamics under each reward's native scale (TensorBoard group-mean reward, EMA-smoothed, $\alpha{=}0.99$). The curves illustrate within-run optimization behavior and stability; their absolute magnitudes are not directly comparable across reward definitions. All variants improve early and remain bounded without sustained collapse.}
\label{fig:reward-curves}
\vspace{-0.17in}
\end{figure}

We then isolate the reward itself. Table~\ref{tab:perfcodebench-leaderboard} includes two simple reward baselines and three component ablations, all trained with the same model, data, seed, and optimizer. Point estimates favor the full reward but cannot establish component necessity. RLVR obtains high correctness as a baseline, but its conditional \textbf{Slow/CRR} rate is $36.6\%$, compared with $15.0\%$ for RLPF. Thus, a substantially larger fraction of RLVR's correct programs fail to improve over the baseline. The runtime-only baseline has the opposite problem: it directly targets speed, but its training signal is sparse and noisy because raw runtime is useful only after correctness and has task-dependent scale.

Figure~\ref{fig:reward-curves} diagnoses within-run behavior rather than comparing reward values. All trajectories rise rapidly and then fluctuate within bounded ranges, without sustained collapse. On its native scale, the full reward improves through most of training but recedes after a late peak, making checkpoint selection important. The ablations stabilize earlier, and the late spike without shaping is transient. Runtime-only training also remains bounded, so its lower value is not evidence of inferior optimization. Because the scales differ and the plot does not report zero-variance rollout groups, we do not compare absolute heights. Cross-variant quality is instead evaluated with the common test metrics in Table~\ref{tab:perfcodebench-leaderboard}.

\subsection{Out-of-Distribution Transfer}
\label{sec:exp-ood}

\begin{table}[t]
\centering
\small
\setlength{\tabcolsep}{4.5pt}
\renewcommand{\arraystretch}{1.1}
\begin{tabular}{lccc}
\toprule
\textbf{Reward} & \textbf{$\Delta$ET (\%)} & \textbf{Win (\%)} & \textbf{$\Delta$MI (\%)} \\
\midrule
\rowcolor{blue!6}
\textsc{RLPF} (full) & \bfseries +3.9 & \bfseries 57.9 & \bfseries +4.3 \\
\midrule
\quad w/o FBR        & -0.3 & 54.5 & \phantom{+}0.0 \\
\quad w/o RBR        & -0.3 & 51.6 & -0.3 \\
\quad w/o shape      & -2.9 & 50.5 & -2.0 \\
\quad runtime only   & +1.1 & 50.9 & +1.0 \\
\bottomrule
\end{tabular}
\vspace{-0.07in}
\caption{\textbf{Out-of-distribution} efficiency on EffiBench-X (five evaluated
languages). Each row is a paired comparison against the base model on the
problems \emph{both} solve: $\Delta$ET / $\Delta$MI are the per-problem
geometric-mean improvements in execution time and memory integral
(higher\,$=$\,better), and Win is the fraction of problems faster than base.
Full RLPF shows the most consistent positive point estimates, while the
evidence for transfer remains modest. Largest point estimates are shown in
\textbf{bold}.}
\label{tab:effibench-x-ood}
\vspace{-0.24in}
\end{table}

Finally, we test whether the learned behavior is tied to the PerfCodeBench harness. We evaluate a five-language EffiBench-X subset with no task overlap with training. All models are decoded with thinking disabled, and each trained model is compared with the same base model under the same protocol. Correctness changes little on EffiBench-X, so the main signal is efficiency on shared solved problems. Full RLPF improves execution time by $3.9\%$ in geometric mean and wins on $57.9\%$ of paired problems. These are the most consistently positive point estimates among the tested reward variants, but the gain is modest and the table does not establish statistical significance. Overall, the results provide preliminary evidence of transfer beyond PerfCodeBench, not proof that every reward component is necessary for out-of-distribution generalization.

\section{Conclusion}
Passing tests is only the first step for systems code. Once a program is correct, the harder question is whether the model can prefer the implementation that uses less time and better matches an optimized reference. This requires feedback that is executable, comparable across tasks, and still useful before the model reaches correctness. RLPF provides such feedback by connecting execution progress with performance improvement. On PerfCodeBench, this changes Qwen3-32B from a model that rarely reaches the measurable regime into one that solves and speeds up many tasks: CRR rises from $11.1\%$ to $54.6\%$, and CGRE rises from $8.1\%$ to $38.6\%$. The same pattern appears in the ablations and, more modestly, in out-of-distribution transfer: the trained model not only passes more tests but also produces faster correct code more often. The next challenge is to move from useful speedups to expert-level implementations. RLPF narrows this gap, but does not close it. Stronger reference signals and better pressure near the expert ceiling are therefore the natural next step.

\section{Limitations}
  
RLPF relies on executable feedback, so its quality depends on the benchmark harness, correctness oracle, and runtime measurements. Repeated evaluation and relative metrics reduce noise, but timing and benchmark test suites remain imperfect for heterogeneous systems tasks~\cite{le2026rethinking}. The method also requires a meaningful performance gap between the baseline and expert reference; weak or unavailable references provide less useful reward. RLPF finds useful speedups more reliably than it matches expert implementations, which often require algorithmic, memory-layout, parallelization, or hardware-specific changes that scalar feedback alone may not reveal. Finally, modest out-of-distribution gains indicate only partial transfer beyond PerfCodeBench.

\bibliography{aaai2027}


\appendix

\section{Dataset Overview}

PerfCodeBench contains $1{,}854$ executable performance tasks. Each task provides a fixed interface, a baseline implementation, an optimized reference implementation, a correctness oracle, and a benchmark harness. The train, validation, and test splits contain $1{,}413$, $135$, and $306$ tasks, respectively. The test split is family-disjoint from training, so evaluation measures whether the learned optimization behavior transfers to held-out task families rather than to near-duplicate prompts.

\begin{table}[t]
\centering
\small
\begin{tabular}{lrrrr}
\toprule
Language & Train & Val & Test & Total \\
\midrule
C++ & 869 & 55 & 163 & 1087 \\
C & 114 & 16 & 32 & 162 \\
CUDA & 93 & 16 & 16 & 125 \\
Go & 112 & 16 & 32 & 160 \\
Java & 128 & 16 & 32 & 176 \\
Python & 97 & 16 & 31 & 144 \\
\midrule
Total & 1413 & 135 & 306 & 1854 \\
\bottomrule
\end{tabular}
\caption{Language distribution of PerfCodeBench.}
\label{tab:app-dataset-language}
\end{table}

The dataset is intentionally heterogeneous. C++ tasks include joins, selection-vector gathers, packed integer decoding, hashing, bitmap filtering, and SIMD-oriented string processing. C, Go, Java, and Python tasks include CSV filtering, dictionary decoding, delta prefix sums, fixed-record serialization, log scanning, JSON field extraction, top-$k$ selection, and group-by histograms. CUDA tasks include reductions, scans, histograms, tiled transpose, and sorting. This diversity is important for RLPF because a useful performance reward must compare candidates across different languages, runtime scales, and failure modes.

\section{Compute Configuration and Cost}

Training is run on one local machine with $8$ NVIDIA A800-SXM4-80GB GPUs. We use Qwen3-32B as the policy model, LoRA with rank $32$ and LoRA alpha $64$, BF16 training, DeepSpeed ZeRO-3 sharding, GRPO group size $G=8$, per-device batch size $1$, gradient accumulation $4$, and maximum completion length $768$. During training, each sampled candidate is executed once by the PerfCodeBench harness. During evaluation, each runtime is measured three times and summarized by the median.

The training cost is dominated by two sources: model-side rollout generation and executable reward evaluation. The latter includes extraction, compilation or loading, execution, oracle checking, and runtime measurement. It varies by language and by whether the generated program reaches the measurable stage. Therefore, wall-clock time is a more informative cost measure than token count alone.

\begin{table}[t]
\centering
\small
\begin{tabular}{lrrr}
\toprule
Run & Tokens & Hours & GPU-hours \\
\midrule
RLPF & 52.6M & 74.6 & 597 \\
RLVR baseline & 51.0M & 69.2 & 553 \\
Runtime baseline & 53.2M & 69.1 & 553 \\
w.o. FBR & 52.8M & 73.1 & 585 \\
w.o. RBR & 52.9M & 71.6 & 573 \\
w.o. shaping & 53.0M & 72.7 & 581 \\
GPT-5.4 reference & 51.4M & 67.9 & 543 \\
\bottomrule
\end{tabular}
\caption{Approximate training cost from trainer logs. GPU-hours are wall-clock hours multiplied by $8$ A800 GPUs.}
\label{tab:app-compute-cost}
\end{table}

\section{Failure Mode Analysis}

PerfCodeBench lets us inspect where a candidate fails in the executable pipeline. This matters because performance feedback is only meaningful after the program is extractable, compiles or loads, runs, and passes the oracle. In the base Qwen3-32B evaluation, only $34$ of $306$ test tasks reach the correct region. Most failures happen earlier: $225$ tasks fail during compilation or execution, and $6$ time out. After RLPF, the number of correct tasks rises to $167$, while compile-or-run failures fall to $90$.

\begin{table}[t]
\centering
\small
\setlength{\tabcolsep}{3.2pt}
\begin{tabular}{@{}lrrrr@{}}
\toprule
Model & Corr. & FBR & RBR & C/R fail \\
\midrule
Qwen3-32B & 34 & 25 & 17 & 225 \\
RLPF & 167 & 142 & 79 & 90 \\
RLVR & 153 & 97 & 50 & 60 \\
Runtime & 116 & 97 & 54 & 97 \\
\bottomrule
\end{tabular}
\caption{Failure-mode summary on the PerfCodeBench test split. Counts are over $306$ tasks. C/R fail denotes compile-or-run failure.}
\label{tab:app-failure-modes}
\end{table}

The remaining failures fall into several patterns. Some candidates still fail because they use the wrong interface, omit required imports, or generate code that is plausible in isolation but invalid as a drop-in implementation. Some candidates run but fail the oracle, especially in parsing and serialization tasks where boundary cases matter. Another important group is correct but slower than the baseline. These cases show why correctness-only reinforcement learning is insufficient: passing tests does not imply that the implementation avoids redundant parsing, extra allocation, unnecessary synchronization, or high-constant-factor library calls.

\section{Reward Baselines and Component Ablations}

The isolated reward experiments separate three regimes. The RLVR baseline gives a correctness-only reward. It solves many tasks, but a large fraction of correct programs remain slower than the baseline. The runtime-only baseline directly targets speed, but it has a sparse and task-dependent signal because runtime is useful only after correctness. The component ablations remove one part of RLPF at a time and show that the full reward is not explained by a single term.

\begin{table}[t]
\centering
\small
\begin{tabular}{lrrr}
\toprule
Variant & Correct & FBR & CGRE \\
\midrule
RLPF & 167 & 142 & 38.6 \\
Runtime reward only & 116 & 97 & 24.7 \\
RLVR correctness only & 153 & 97 & 29.0 \\
w.o. FBR & 124 & 98 & 25.4 \\
w.o. RBR & 122 & 95 & 25.1 \\
w.o. shaping & 121 & 95 & 24.2 \\
\bottomrule
\end{tabular}
\caption{Isolated reward effects on the PerfCodeBench test split. Counts are over $306$ tasks, and CGRE is reported as a percentage.}
\label{tab:app-reward-components}
\end{table}

The terms play different roles. Failure-mode shaping gives GRPO an ordering before correctness. The FBR bonus rewards the first useful step beyond the baseline. The RBR bonus keeps pressure near the expert reference. Together, these terms make the reward useful both before and after correctness, which is the main difference between RLPF and simpler reward designs.

\section{Representative Case Studies}

Several test-set examples illustrate the behavior learned after training. On \texttt{openmp\_sum\_atomic\_v900\_j}, the base model fails before producing a valid measurable program. RLPF produces a correct implementation that is about $101\times$ faster than the baseline. This task family rewards reducing synchronization-heavy accumulation and exposing more parallel work.

On \texttt{java\_bitset\_and\_popcount\_v3200\_106}, the base model again fails in the executable pipeline. RLPF reaches a correct solution that is about $51\times$ faster than the baseline and matches the reference threshold. The improvement comes from selecting a more suitable low-level primitive and avoiding per-element overhead.

RLPF also improves smaller data-processing cases. On \texttt{python\_json\_field\_checksum\_v2700\_029}, the RLVR baseline produces a correct program that is slightly slower than the baseline, while RLPF produces a correct faster program and reaches the reference threshold. On \texttt{go\_groupby\_histogram\_checksum\_v3300\_124}, RLVR is again correct but slower, whereas RLPF beats the baseline. These examples show the difference between learning to pass the oracle and learning to prefer leaner correct code.

\begin{table*}[t]
\centering
\small
\begin{tabular}{lrrrr}
\toprule
Task & Base & RLPF & Speedup & Main pattern \\
\midrule
\texttt{openmp\_sum\_atomic\_v900\_j} & fail & correct & $101.1\times$ & lower synchronization cost \\
\texttt{java\_bitset\_and\_popcount\_v3200\_106} & fail & correct & $50.9\times$ & better low-level primitive \\
\texttt{dictionary\_decode\_checksum\_v1200\_b} & fail & correct & $21.8\times$ & reference-level decoding path \\
\texttt{python\_json\_field\_checksum\_v2700\_029} & fail & correct & $1.3\times$ & less parsing overhead \\
\texttt{go\_groupby\_histogram\_checksum\_v3300\_124} & fail & correct & $1.1\times$ & leaner aggregation path \\
\bottomrule
\end{tabular}
\caption{Representative test-set cases improved by RLPF. Speedup is measured against the task baseline.}
\label{tab:app-case-studies}
\end{table*}

\section{Model-Generated Performance References}

Expert references provide the cleanest performance target, but they require strong benchmark construction. We therefore also test a weaker supervision setting where a correct GPT-5.4 candidate supplies the reference runtime. This setting approximates a practical distillation scenario: a strong model can provide performance supervision on tasks it solves correctly, even when a curated expert implementation is unavailable.

The GPT-5.4-reference variant reaches $50.3\%$ CRR and $36.5$ CGRE, close to full RLPF but still weaker on correctness and faster-than-baseline coverage. This suggests that model-generated references are useful but less uniform than expert references. They only cover tasks solved correctly by the teacher, and the teacher may still miss task-specific optimizations that the benchmark reference captures.

\section{Metric Decomposition}

The main paper reports CRR, FBR, RBR, and CGRE because they describe different parts of the optimization pipeline. CRR measures whether a model can produce a correct runnable implementation. FBR measures whether the correct implementation is useful relative to the baseline. RBR measures whether it reaches the expert reference. CGRE gives a continuous view of progress between the baseline and reference.

Two derived quantities are useful for diagnosis. The first is \emph{Slow} $=$ CRR $-$ FBR, the share of programs that are correct but not faster than the baseline. The second is \emph{Gap} $=$ FBR $-$ RBR, the share of programs that beat the baseline but still do not reach the reference. RLPF mainly improves the transition from correctness to faster-than-baseline code. Its remaining gap is closer to the reference boundary, where task-specific optimization choices still matter.

\begin{table}[t]
\centering
\small
\setlength{\tabcolsep}{4.0pt}
\begin{tabular}{lrrrr}
\toprule
Model & CRR & FBR & Slow & Gap \\
\midrule
Qwen3-32B & 11.1 & 8.2 & 2.9 & 2.6 \\
RLVR & 50.0 & 31.7 & 18.3 & 15.4 \\
Runtime & 37.9 & 31.7 & 6.2 & 14.1 \\
RLPF & 54.6 & 46.4 & 8.2 & 20.6 \\
\bottomrule
\end{tabular}
\caption{Metric decomposition on the PerfCodeBench test split. All values are percentages. Slow denotes correct but not faster than the baseline; Gap denotes faster than baseline but not reference-level.}
\label{tab:app-metric-decomposition}
\end{table}

This decomposition also explains why a single correctness score is not enough. RLVR reaches a high CRR, but many of its correct solutions remain slow. Runtime-only training has a lower Slow value, but it reaches fewer correct programs overall. RLPF keeps the correctness gain while increasing the number of faster-than-baseline programs, which is the desired regime for performance-oriented code generation.

\section{Training Dynamics}

The reward curves are plotted on each reward's native scale, so their absolute heights should not be compared across reward definitions. They are still useful for understanding stability. All variants rise early and remain bounded. The full reward reaches its highest EMA value around step $1552$ and then declines mildly by the final checkpoint. This is why the main experiments use EMA-peak model selection rather than always taking the last checkpoint.

\begin{table}[t]
\centering
\small
\setlength{\tabcolsep}{5.0pt}
\begin{tabular}{lrr}
\toprule
Reward & Peak step & Final EMA \\
\midrule
Full RLPF & 1552 & 0.223 \\
w.o. FBR & 1556 & 0.186 \\
w.o. RBR & 1166 & 0.170 \\
w.o. shaping & 1677 & 0.198 \\
\bottomrule
\end{tabular}
\caption{Training-curve summary from exported TensorBoard EMA values ($\alpha=0.99$). Peak step is selected on the EMA curve.}
\label{tab:app-training-dynamics}
\end{table}

The curve shapes also support the reward design. Removing RBR tends to peak earlier, suggesting that the reward gives less pressure near the expert target. Removing shaping can still learn, but its signal is concentrated after correctness and its late improvement is less stable. The full reward has the strongest late-stage signal because it combines failure-stage progress, baseline improvement, and reference matching.

\section{Additional Case Study Analysis}

The case studies in Table~\ref{tab:app-case-studies} focus on large improvements. We add a second table to show a broader set of behaviors. These examples separate three cases: conversions from failure to fast correct code, conversions from correct-but-slow to faster code, and cases where the model beats the baseline but still falls short of the expert reference.

\begin{table*}[t]
\centering
\small
\setlength{\tabcolsep}{4.0pt}
\begin{tabular}{llrrl}
\toprule
Task & Before RLPF & RLPF speedup & CGRE & Interpretation \\
\midrule
\texttt{openmp\_sum\_atomic\_v900\_j} & fail & $101.1\times$ & 0.99 & removes a severe synchronization bottleneck \\
\texttt{java\_bitset\_and\_popcount\_v3200\_106} & fail & $50.9\times$ & 1.00 & switches to a more suitable bit-level primitive \\
\texttt{dictionary\_decode\_checksum\_v1200\_b} & fail & $21.8\times$ & 1.00 & reaches the intended decoding strategy \\
\texttt{packed12\_decode\_filter\_sum\_lowthr} & RLVR slow & $1.43\times$ & 0.54 & turns a correct solution into a useful speedup \\
\texttt{python\_json\_field\_checksum\_v2700\_029} & RLVR slow & $1.27\times$ & 1.00 & reduces repeated parsing overhead \\
\texttt{go\_groupby\_histogram\_checksum\_v3300\_124} & RLVR slow & $1.14\times$ & 0.63 & improves allocation and aggregation behavior \\
\texttt{sorted\_interval\_lookup\_sum\_v1200\_c} & RLVR slow & $1.01\times$ & 0.01 & barely beats baseline; still far from reference \\
\bottomrule
\end{tabular}
\caption{Additional test-set cases illustrating different improvement regimes. ``RLVR slow'' means the correctness-only baseline passed the oracle but did not beat the baseline runtime.}
\label{tab:app-additional-cases}
\end{table*}

These cases show that RLPF is not only finding rare large speedups. It also changes the preference among correct programs. In parsing and aggregation tasks, the gains can be modest because the baseline is already close to a reasonable implementation. These small gains are still important: they show that the reward can distinguish correct programs by runtime rather than treating them as equivalent.

\section{Remaining Failure Patterns}

RLPF reduces compile and run failures, but it does not remove them. The remaining failures are concentrated in four patterns. First, some generations still violate the required interface, especially when the prompt contains a complete source file but the model emits only a function body. Second, some candidates use unavailable libraries or language features that are not accepted by the harness. Third, some programs pass simple-looking logic but fail boundary cases in parsing, serialization, and integer overflow. Fourth, some hardware-oriented or parallel kernels compile but still miss the intended low-level optimization, such as memory locality, synchronization reduction, or layout-aware access.

\begin{table}[t]
\centering
\small
\setlength{\tabcolsep}{4.0pt}
\begin{tabular}{ll}
\toprule
Pattern & Typical cause \\
\midrule
Interface error & missing entrypoint or wrong signature \\
Build error & unavailable include, package, or flag \\
Oracle failure & boundary case or numeric mismatch \\
Timeout & algorithmic or synchronization bottleneck \\
Slow correct & extra allocation or high constant factor \\
\bottomrule
\end{tabular}
\caption{Qualitative failure patterns observed in test-set traces.}
\label{tab:app-failure-pattern-taxonomy}
\end{table}

This analysis motivates two future directions. The first is better format control before execution, which would reduce avoidable compile failures. The second is more specialized performance supervision for low-level optimization domains, where correctness and performance depend on hardware-aware choices that are hard to learn from sparse successes.

\section{Transition Analysis}

Another way to read the results is to compare task outcomes before and after RLPF. Relative to the base model, RLPF converts $135$ previously incorrect tasks into correct ones, $117$ tasks into faster-than-baseline solutions, and $65$ tasks into reference-level solutions. Relative to RLVR, the gains are more focused: RLPF converts $58$ additional tasks into faster-than-baseline solutions and $40$ additional tasks into reference-level solutions. This supports the main claim that RLPF does not only improve correctness. It changes the ranking among correct programs.

\begin{table}[t]
\centering
\small
\setlength{\tabcolsep}{3.5pt}
\begin{tabular}{lrrrr}
\toprule
Reference & Fail$\to$Corr. & NotFast$\to$FBR & NotRef$\to$RBR & Lost Corr. \\
\midrule
Base & 135 & 117 & 65 & 2 \\
RLVR & 34 & 58 & 40 & 20 \\
Runtime & 59 & 54 & 37 & 8 \\
\bottomrule
\end{tabular}
\caption{Outcome transitions from each reference model to RLPF on the same $306$ test tasks. ``Lost Corr.'' counts tasks solved by the reference model but not by RLPF.}
\label{tab:app-transition-analysis}
\end{table}

The transition table also shows a tradeoff. RLPF is not a strict superset of RLVR. It loses correctness on $20$ tasks that RLVR solves, while gaining correctness on $34$ tasks that RLVR misses. This is expected because the policy is optimized for a different ordering: it must keep correctness, but it also receives pressure to search for faster implementations. The net result is a model that solves slightly more tasks and, more importantly, produces many more useful correct programs.

\section{Speedup Distribution}

The average speedup of faster-than-baseline RLPF outputs is high, but the distribution is skewed. A small number of tasks expose very large gains, while many tasks provide modest but real improvements. The median speedup among faster RLPF solutions is $4.77\times$, and $68$ tasks exceed $5\times$. Only $5$ tasks exceed $50\times$. This suggests that the benchmark contains both low-level bottlenecks with large optimization headroom and tighter data-processing tasks where the best realistic gain is smaller.

\begin{table}[t]
\centering
\small
\setlength{\tabcolsep}{5.0pt}
\begin{tabular}{lr}
\toprule
Speedup threshold & RLPF tasks \\
\midrule
$\geq 1.05\times$ & 115 \\
$\geq 1.20\times$ & 112 \\
$\geq 1.50\times$ & 106 \\
$\geq 2.00\times$ & 97 \\
$\geq 5.00\times$ & 68 \\
$\geq 10.0\times$ & 45 \\
$\geq 20.0\times$ & 24 \\
$\geq 50.0\times$ & 5 \\
\bottomrule
\end{tabular}
\caption{Speedup distribution for the $142$ RLPF outputs that beat the baseline.}
\label{tab:app-speedup-distribution}
\end{table}

This distribution is useful for interpreting CGRE. A small speedup can have high CGRE when the expert reference is also close to the baseline. Conversely, a large absolute speedup can still have moderate CGRE if the reference is much faster. For this reason, the paper reports both threshold metrics and the continuous relative-efficiency metric.

\section{Task-Family Contributions}

RLPF gains are concentrated in task families with clear implementation bottlenecks. Table~\ref{tab:app-family-contrib} lists representative families with at least three test tasks. Some families are nearly saturated: \texttt{topk\_ordered\_sum}, \texttt{dictionary\_decode\_checksum}, and \texttt{dense\_groupby\_sum} have high CGRE once solved. Other families, such as \texttt{sorted\_interval\_lookup\_sum}, produce many faster-than-baseline solutions but low CGRE, meaning that RLPF improves over the baseline without reaching the expert design.

\begin{table*}[t]
\centering
\small
\setlength{\tabcolsep}{4.4pt}
\begin{tabular}{lrrrr}
\toprule
Task family & Tasks & Correct & FBR & Mean CGRE \\
\midrule
\texttt{sorted\_interval\_lookup\_sum} & 19 & 19 & 18 & 0.01 \\
\texttt{java\_bitset\_and\_popcount} & 16 & 16 & 16 & 1.00 \\
\texttt{java\_fixed\_record\_serialize} & 16 & 16 & 15 & 0.91 \\
\texttt{openmp\_sum\_atomic} & 16 & 14 & 14 & 0.77 \\
\texttt{python\_json\_field\_checksum} & 16 & 16 & 11 & 0.65 \\
\texttt{topk\_ordered\_sum} & 11 & 11 & 11 & 1.00 \\
\texttt{dictionary\_decode\_checksum} & 11 & 11 & 11 & 1.00 \\
\texttt{dense\_groupby\_sum} & 11 & 11 & 11 & 0.99 \\
\texttt{csr\_row\_sum\_checksum} & 11 & 11 & 11 & 0.99 \\
\texttt{go\_groupby\_histogram\_checksum} & 16 & 16 & 5 & 0.19 \\
\bottomrule
\end{tabular}
\caption{Representative task-family contributions for RLPF. Families are grouped by normalized task identifier. Mean CGRE is averaged over all tasks in the family, including failures.}
\label{tab:app-family-contrib}
\end{table*}

The table highlights two different success modes. In families such as bitset popcount and dictionary decoding, a single structural choice often moves the program close to the reference. In families such as interval lookup and Go group-by histograms, RLPF often improves the code but still trails the expert. These latter families are useful stress tests because they require fine-grained choices rather than one obvious replacement.

\section{Failure-Stage Counts}

The executable harness records whether a candidate is extracted, compiled, run, and judged correct. In the base model, most failures are compilation failures. RLPF reduces this number from $266$ to $131$, while increasing correct programs from $34$ to $167$. Run-stage failures remain small in absolute count. This indicates that most of the improvement comes from producing code that matches the expected replacement format and survives compilation, after which the reward can provide performance signal.

\begin{table}[t]
\centering
\small
\setlength{\tabcolsep}{5.0pt}
\begin{tabular}{lrr}
\toprule
Outcome stage & Base & RLPF \\
\midrule
Correct & 34 & 167 \\
Compile failure & 266 & 131 \\
Run failure & 6 & 8 \\
\bottomrule
\end{tabular}
\caption{Failure-stage counts from the test-set execution traces. Counts are mutually exclusive and sum to $306$.}
\label{tab:app-failure-stage-counts}
\end{table}

The slight increase in run failures is not a major regression by itself. As more generations compile, more candidates reach later execution stages where runtime errors and oracle failures can be observed. This is the intended behavior of staged execution feedback: earlier-stage failures should move forward in the pipeline, even if not all of them become correct immediately.

\section{Prompt Templates}

Training and in-house evaluation use the same task prompt template. The only difference in the OpenAI-based evaluation path is the API wrapper and its strict JSON schema; the underlying task prompt is the same. We include the full templates here for reproducibility.

\begin{listing}[t]
\caption{System prompt used during rollout generation and evaluation.}
\label{lst:app-system-prompt}
\begin{lstlisting}
You are a performance engineer. Your job is to rewrite the given source file so that it runs as fast as possible while preserving correctness. Do not change the function signature or entrypoint used by the harness. Output JSON only, with keys 'summary' and 'solution_source'. The solution_source must be a complete, compilable replacement file.
\end{lstlisting}
\end{listing}

\begin{listing}[t]
\caption{User prompt template shared by training and in-house evaluation.}
\label{lst:app-user-prompt}
\begin{lstlisting}
Optimize the following implementation for performance while preserving correctness.
Return a full, complete, compilable replacement for <source_name>.
Do not change the externally required function signature or entrypoint.

Task ID: <task_id>
Title: <title>
Goal: <goal>
Metric: <metric>
Correctness rule: <correctness_rule>
Allowed external includes: <comma-separated list>

Interface / task contract:
<optional interface text>

Current baseline <source_name>:
<baseline source>
\end{lstlisting}
\end{listing}

\begin{listing}[t]
\caption{OpenAI evaluation wrapper. The prompt is the same as above, but the request also enforces a strict JSON schema.}
\label{lst:app-json-schema}
\begin{lstlisting}
{
  "type": "object",
  "additionalProperties": false,
  "properties": {
    "summary": {"type": "string"},
    "solution_source": {"type": "string"}
  },
  "required": ["summary", "solution_source"]
}
\end{lstlisting}
\end{listing}

\end{document}